\documentclass[11pt,a4paper]{article}
\usepackage{times,latexsym}
\usepackage{url}
\usepackage[T1]{fontenc}

\usepackage[acceptedWithA]{tacl2021v1}
\usepackage{tacl2021v1}

\usepackage{xspace,mfirstuc,tabulary}

\usepackage{graphicx}
\usepackage{tabularx}
\usepackage{soul}
\usepackage{epstopdf}
\usepackage{todonotes}
\usepackage[utf8]{inputenc}
\usepackage{svg}
\usepackage{hyperref}
\usepackage{xstring}
\usepackage{color}
\usepackage{booktabs}
\usepackage{pifont}
\usepackage{amsmath}
\usepackage{multirow}
\usepackage{xspace}
\usepackage{subcaption}
\usepackage{epsfig, endnotes}
\usepackage{array}
\usepackage{float}
\usepackage{mfirstuc, tabulary}
\usepackage{paralist}
\usepackage{csquotes}
\usepackage{enumitem}
\usepackage{wasysym}
\usepackage{amssymb}
\usepackage{pgf-pie}
\usepackage{hyperref} 
\usepackage{nameref}
\usepackage{makecell}
\usepackage{cuted}
\usepackage{xcolor}
\usepackage{enumitem}
\usepackage{subcaption,siunitx,booktabs}

\newcommand{\dataset}[0]{\texttt{PASTA}}

\newcommand{\storyrev}[0]{Story Revision for Counterfactual States}
\newcommand{\statechange}[0]{State Change Generation}
\newcommand{\mcq}[0]{Story State Inference}

\newif\iftaclinstructions
\taclinstructionsfalse 
\iftaclinstructions
\renewcommand{\confidential}{}
\renewcommand{\anonsubtext}{(No author info supplied here, for consistency with
TACL-submission anonymization requirements)}
\newcommand{\instr}
\fi

\iftaclpubformat 

\else

\fi

\title{\dataset{}: A Dataset for Modeling {PA}rticipant {STA}tes in Narratives}

\author{
  Sayontan Ghosh$^\diamondsuit$\hspace{1.5em} 
  Mahnaz Koupaee$^\diamondsuit$\hspace{1.5em}
  Isabella Chen$^{\diamondsuit\dagger}$ \\
  \textbf{Francis Ferraro}$^\spadesuit$\hspace{1.5em}
  \textbf{Nathanael Chambers}$^\clubsuit$\hspace{1.5em}
  \textbf{Niranjan Balasubramanian}$^\diamondsuit$ 
  \\ \ \\
  {$^\diamondsuit$Stony Brook University}\hspace{1em}
  $^\spadesuit$University of Maryland, Baltimore County\hspace{1em}
  $^\clubsuit$United States Naval Academy \hspace{1em}
  \\ 
  \texttt{\{sagghosh, mkoupaee, niranjan\}@cs.stonybrook.edu}
  \\ \texttt{isabellachenusa@gmail.com}\hspace{1em}
  \texttt{ferraro@umbc.edu}\hspace{1em}
  \texttt{nchamber@usna.edu}
}

\date{}

\begin{document}
\maketitle
\begin{abstract}
The events in a narrative are understood as a coherent whole via the underlying states of their participants. Often, these participant states are not explicitly mentioned, instead left to be inferred by the reader.
A model that understands narratives should likewise infer these implicit states, and even reason about the impact of changes to these states on the narrative. 
To facilitate this goal, we introduce a new crowdsourced English-language, \emph{Pa}rticipant \emph{Sta}tes dataset, \dataset{}. This dataset contains inferable participant states; a counterfactual perturbation to each state; and the changes to the story that would be necessary if the counterfactual were true. We introduce three state-based reasoning tasks that test for the ability to infer when a state is entailed by a story, to revise a story conditioned on a counterfactual state, and to explain the most likely state change given a revised story. Experiments show that today's LLMs can reason about states to some degree, but there is large room for improvement, especially in problems requiring access and ability to reason with diverse types of knowledge (e.g. physical, numerical, factual).\footnote{Code and dataset are available at \url{https://github.com/StonyBrookNLP/pasta} \par $^\dagger$Work done during internship at Stony Brook University.}

\end{abstract}
\section{Introduction}
\begin{figure*}[!ht]
    \includegraphics[width=\textwidth]{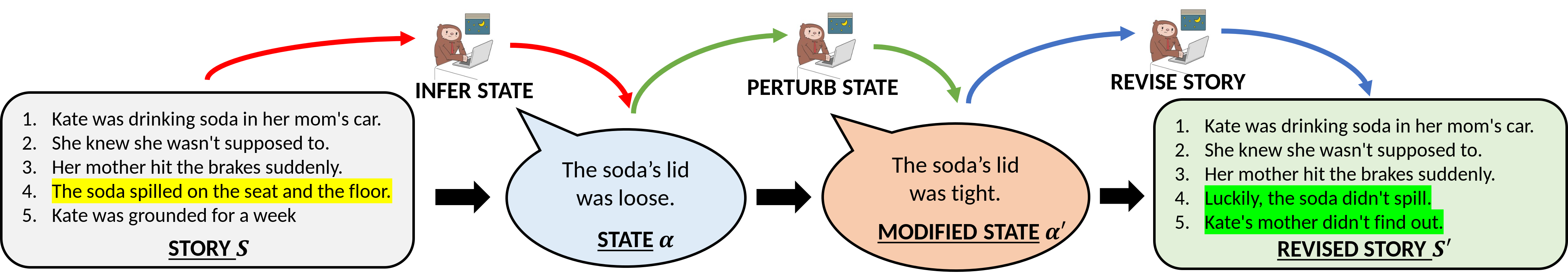}
    \caption{\small{For a given story $S$, \textbf{\dataset{}} provides an (unstated) inferred state $\alpha$, a minimal set of justification sentences (\hl{yellow}   highlight), a counterfactual state $\alpha'$, and a revised story $S'$, such that $\alpha'$ can be inferred from it.}
    }
    \label{Fig:pasta_example}
\end{figure*}

Understanding narrative text requires forming a coherent representation of the scenario, including filling in details that are unstated in the text. One type of detail that is usually not mentioned are the states of its participants\footnote{We define participants to include both animate entities and inanimate objects in the narratives.} (e.g., "she unlocked the door" implies the possession state that "she has a key").
%
The reader easily infers these \emph{implicit} states and their causal relationships with the narrative's \emph{explicit} events, creating a detailed mental picture of the described world that is only partially observable from the text.
Many cognitive theories have been proposed to capture aspects of this in their representations, such as scripts~\cite{schank1975scripts}, frames~\cite{fillmore1985frames}, and state/time formalisms~\cite{galton1990critical}.
Without committing to any one particular formal theory, this paper adds a theory-agnostic resource 
to test such theories by listing implicitly assumed participant states in simple narratives.

Consider the story in Figure~\ref{Fig:pasta_example} from the ROCStories corpus \cite{mostafazadeh2016corpus}.
Humans create a detailed mental representation of this spilled-soda scenario by inferring its commonsense states. %
In this story, using our commonsense knowledge about emotions and habituals, we can infer from the first two lines that \underline{Kate's mother liked keeping her car clean} (a state about \emph{Kate's mother}). Similarly, based on our physical commonsense of a lid, i.e., that lids prevent spilling, we can also assert from the spill that \underline{the soda's lid was loose} (a state about the \emph{soda}). 
We can also reason about the likely change to the story due to a counterfactual state, i.e., if \underline{the soda's lid was tight}, then most likely the soda wouldn't spill.
To the best of our knowledge, no such resource exists that captures this kind of participant state knowledge.
To capture this type of commonsense knowledge needed to understand and reason about participant states in narratives,
we introduce \dataset, a crowd-sourced dataset in English.
%
%
As shown in Figure~\ref{Fig:pasta_example}, for a given story $S$, \dataset{} provides a participant state $\alpha$ that is likely to be inferred from $S$, a perturbation state $\alpha'$ that is counterfactual to $S$ along with the minimal changes to $S$ that are required to make $\alpha'$ likely to be inferred from the revised story $S'$.
\dataset{} includes $10,743$ instances of these story/state/counterfactual/revision tuples.
%
With this new dataset, we hope to enable models to make the kinds of state-based inferences that move beyond surface text understanding and lead to deeper reasoning.
To this end, we describe three new state-based reasoning challenges with \dataset{} which are illustrated in Figure~\ref{Fig:stae_tasks}.

The first is \textbf{\mcq}: given a story and an inferred participant state, predict if the state is likely to be inferred from a given set of sentences in the context of the story.
We formulate this as a binary classification task, and we create contrastive examples for training and evaluation purposes to guard against artifact-based reasoning. %
This can be seen as a form of textual entailment, a capability useful for applications such as question answering~\cite{harabagiu-hickl-2006-methods, trivedi-etal-2019-repurposing}, claim verification~\cite{yin-roth-2018-twowingos, hanselowski-etal-2018-ukp}, etc.

The other two challenge tasks are generative. 
The second task, \textbf{\storyrev{}}, measures the ability to reason about counterfactuals. Given a story and a counterfactual state (i.e., a state that is not consistent with the story), the task is to revise the story such that the counterfactual state is now likely to be inferred from it. These types of counterfactual revisions serve as a test of reasoning~\cite{Qin2019CounterfactualSR} and can support interactive story generation tasks~\cite{goldfarb-tarrant-etal-2019-plan, brahman-etal-2020-cue}.
%
The third task, \textbf{\statechange{}}, requires the model to take a story and its perturbed version as input and then generate the two corresponding states (e.g., `lid was loose' and `lid was tight') that explain the differences in the way they unfold.
From an application perspective, generating the underlying states that account for the differences between two narratives can assist with fake news detection using reliable sources \cite{figueira2017current, cardoso2019can, ghadiri2022automated} and information fact checking \cite{brandtzaeg2018journalists}.

%
These three challenge tasks require a unique combination of commonsense abilities, thus helping to evaluate models on reasoning and knowledge capacity. These tasks require not only basic entailment ability, but also knowledge (numerical, factual, physical, etc.) and broader narrative understanding. Having just one of these abilities will not suffice.
%
To evaluate current models for these capabilities, we benchmark the LLMs T5~\cite{raffel2020exploring}, BERT~\cite{devlin-etal-2019-bert}, RoBERTa~\cite{Liu2019RoBERTaAR} and GPT3~\cite{brown2020language}. For the generative tasks, we evaluate model performance through extensive human and automatic evaluations.
The results show that, though these models can reason about states to some degree, there is substantial room for improvement on all tasks, suggesting avenues for future research.

\section{Related Work}
%
There are many formal theories on mental states and reasoning. The seminal work by \citet{schank1975scripts} introduced scripts as a way to structure knowledge about stereotypical event sequences with their participants. Frames \cite{fillmore1985frames} and theories of time \cite{galton1990critical} provide related views. This paper does not commit to a formal theory, instead providing a challenge dataset to test aspects of them.
Statistical work on events \cite{chambers2008unsupervised, chambers2009unsupervised, balasubramanian2013generating,ferraro-2016-upf,sha2016joint} curated event knowledge in an unsupervised manner from large text corpora. This paper augments their view with \emph{state-based} knowledge about event participants.
%
\par
More recent work by \citet{speer2017conceptnet},  \citet{sap2019atomic} and \citet{Hwang2021COMETATOMIC2O} capture everyday inferential knowledge associated with an action performed by someone. This knowledge is organized through a fixed set of relationship classes between action and inferences. A subset of these classes is about participant mental states, but this commonsense knowledge is non-contextual in nature. In contrast, our work requires inferring commonsense knowledge about participant states in the context of coherent narratives. 
\par
Most similar to our work is the {TIME-TRAVEL} dataset by \citet{Qin2019CounterfactualSR}. It includes the Counterfactual Story Rewriting task to edit a short story based on a counterfactual context.
They insert an explicit counterfactual at a fixed position ($2^{nd}$ sentence) in the story, and the revision task is then conditioned on this observed change. It is a language modeling generation task.
In contrast, our work introduces \emph{unobserved} counterfactual outside of the story's text, and the revised story must be generated with deeper state-based reasoning.
This introduces additional complexity for the revision task.
Also, {TIME-TRAVEL} requires the revisions to be restricted to the story ending, which cannot be assumed in our setting. Our states can be inferred from any part of the story.
\par
\citet{bhagavatula2019abductive} proposed tasks that predict a plausible hypothesis for two given observations, and curate a dataset for the same. Their work mainly focuses on \textit{what happened in-between?} type of inferences.
\citet{mostafazadeh-etal-2020-glucose} introduced the {GLUCOSE} dataset, which focuses on several types of causal knowledge that is required to explain a causal event in narrative text.
Neither of these focuses entirely on implicit states (some {GLUCOSE} annotations are relevant, but not directly so), and neither addresses story revision in the face of counterfactual changes.
\par
Recent work on understanding entity states has mostly focused on tracking entity state change in text. 
\citet{Dalvi2018TrackingSC} introduced {PROPARA} which captures physical state changes (creation, destruction, and movement), \citet{Bosselut2018SimulatingAD} proposed the task of tracking ingredients in cooking recipes, and \citet{rashkin-etal-2018-modeling} tracks the emotional reactions and motivations of characters in simple stories, for a fixed/small set of attributes.
\par
\citet{tandon-etal-2019-wiqa} introduced the {WIQA} dataset for analyzing the effect of perturbing a process described by a procedural text on the elements (entities, events, etc.) of the text, as an influence graph of the process. However, the influence graph was assumed to have a fixed causal structure. It captured a very limited set of cause-effect relationships obtained as a result of analyzing perturbations that either accelerated or decelerated the main outcome of the process.
\citet{tandon-etal-2020-dataset} introduced a dataset for tracking state changes in procedural text as a set of state change tuples of \textit{entity, attribute, before-state}, and \textit{after-state} for each step of the process. The elements of the tuples were in free-form text instead of belonging to a set of pre-defined categories. 
\par
Our work differs from the above in several key ways: (i) participant states are unstated (ii) participant state inferences do not depend on sentence ordering assumptions, (iii) state perturbations affect the entire discourse of the narrative, and (iv) captures how the participant states change between an original and a perturbed narrative.

\section{\dataset{}: {PA}rticipant {STA}tes}

\dataset\ is a dataset of story pairs ($S$, $S'$) where each story $S$ has a revised version of itself, $S'$, that hinges on a particular state that was changed in its revision. The story pairs thus have corresponding state pairs ($\alpha$, $\alpha'$), containing an original state $\alpha$ and its counterfactual $\alpha'$. Refer to Figure~\ref{Fig:pasta_example}. These story/state pairs allow us to analyze unique narrative challenges. We can test if a model can identify whether a given participant state is consistent with it. We can ask what would happen if an assumed story state is no longer true. We can also ask if a model can identify what/how a state changes between two similar but different stories.
This section describes \dataset{}, the crowd-sourcing process that created it, quality control details, and its basic statistics.
%
\subsection{Data annotation}
To create the \dataset{} dataset, we use stories from the extended ROCStories \cite{mostafazadeh2016corpus} corpus for annotation by crowd workers. 
ROCStories narratives describe a rich set of causal and temporal commonsense relations between daily events, and its stories are short enough that the world described by them are self-contained. They thus are a good fit for testing state inferences.

Figure~\ref{Fig:pasta_example} illustrates the process followed to collect responses from the crowd workers.\footnote{The project was reviewed and approved by the local IRB for human subjects research.} %
%
The annotation process has four main steps: 
\begin{enumerate}[noitemsep, nolistsep]
    \item \textbf{Infer a participant state}: For a story $S$, the annotator \underline{infers} a participant (or object) state $\alpha$ that is likely to be true at some point in $S$; $\alpha$ is a free-form sentence. Most stories have several inferrable states, so the annotator may identify whatever jumps out to them the most.
    \item \textbf{Select minimal justification sentences:}
    For the inferred $\alpha$, the annotator \underline{selects} the minimal set of sentences $J^S_{\alpha}$ in $S$, that they used to infer $\alpha$ from $S$. 
    \item \textbf{Perturb the state:}
    The annotator \underline{perturbs} $\alpha$ to create $\alpha'$ such that $\alpha'$ is very unlikely to be true for the story $S$. $\alpha'$ is also a free-form sentence.
    \item \textbf{Revise the story:}
    The annotator \underline{revises} $S$ into $S'$, so that $\alpha'$ can be inferred from $S'$ but $\alpha$ is unlikely to be inferred from $S'$. The annotator is instructed to make minimal revisions in order to avoid creating $S'$ with other narrative side-effects.
\end{enumerate}
%
%
We provide detailed instructions about how to infer a state, and these are repeated not just in the instructions and examples, but also in the actual form they fill out. 
The inferred state must be a property or attribute of a participant or object (e.g., \emph{she was angry} or \emph{the rock is heavy}); it must not be an action (e.g., \emph{Susan is running} or \emph{Jake cooks food}); and it must not be explicitly stated in the story. %
These constraints assure that the states are not readily available from the story text, and must be inferred by reasoning and world knowledge. The next section describes how we monitored the workers and mitigated improper responses.


\subsection{Quality Control}\label{subsec:quality_control}
For crowdsourcing the data collection, we used the Amazon-MTurk platform (AMT). Each story was provided to three different crowd-workers for annotation. 
We priced the HIT at $\$0.35$ based on initial worker response times and interest gleaned from multiple pilot runs.
For filtering out noisy data from the collected responses, we follow a two-stage filtering process.

\paragraph{Stage 1:}
We only allowed workers with a long history of consistent performance who satisfy the following criteria:
\begin{enumerate}[noitemsep, nolistsep]
    \item have responded to at least 5000 HITs
    \item have at least 98\% accuracy on their past HITs
    \item must reside in USA or Canada; this helps to prevent language-based artifacts
\end{enumerate}
Although the above is strict, we still observed responses that did not follow the instructions. One difficulty was how workers wrote their revised stories. Even minor changes to the original story can render it logically inconsistent, so care is needed to ensure the counterfactual is inferrable while still maintaining coherence.
Other annotation errors were `states' describing actions, states directly mentioned in the story, and non-entailed states.

\vspace{-1pt}

\paragraph{Stage 2:} 
Despite the above errors, we received excellent responses with clear states and interesting revised stories. This gave us confidence that the task is achievable, but it just needed expert crowd workers.
To this end, we performed an "expert review" of the responses to identify "proficient workers": workers who can perform the task with a high degree of correctness. Our expert reviewers are two student researchers who work in the field of common-sense reasoning and NLP in general. Stage 1 resulted in a total of $9656$ responses from 136 workers. The experts evaluated a subset of these to identify proficient workers by using the process described below:

\vspace{-1pt}

%
\begin{enumerate}[noitemsep,nolistsep]
    \item For each worker, we manually evaluated their performance on a random sample of their responses. 
    \item The number of evaluated responses for each worker was decided by the formula below. If the $i^{th}$ worker provides $n_i$ responses, then the minimum number of their responses, $e_i$, that needs to be expert-reviewed to evaluate their proficiency is given by: 
    \[ e_i = \begin{cases} 
          0.3*n_i & n_i < 100 \\
          0.2*n_i + 10 & 100\leq n_i < 150 \\
         40 & 150 \leq n_i 
       \end{cases}
    \]
    \item Each evaluated response was categorized as correct or reject. A response was rejected if there was an error in any of the four steps of the annotation process. A response is correct if all the components of the annotation adheres to the instructions.
    \item A worker was identified as proficient if they submitted $\geq50$ responses with a rejection rate $\leq20\%$.
    After identifying proficient workers, all other responses from proficient workers were then auto-accepted. We also kept the smaller number of non-reject responses that our experts labeled from non-proficient workers.
\end{enumerate}

With this process, we identified 28 workers who were proficient. We accepted all of their annotations totaling $\sim6,000$. To this we added the annotations the experts accepted in the review, which added another $360$ high quality instances. We then ran a second round of data collection using only the proficient workers. We added this to the high quality instances from the first round to form our full \dataset{} dataset.


The responses in the pool of expert-reviewed responses were used to create the test set of the data. We also made sure that there is no story overlap in the train, validation, and test sets.

\subsection{Dataset Statistics}
\dataset{} includes a total of $10,743$ ($8476$ train, $1350$ validation, and $917$ test) 4-tuples. Each 4-tuple is a story $S$, an associated inferred state $\alpha$, counterfactual state $\alpha'$, and a revised story $S'$.
Annotators almost always changed the justification sentences of the inferred state in order to revise the story. Instructions to make minimal changes to the revised story results in a high degree of similarity between the original and revised stories. On average $1.5$ out of $5$ story sentences are changed to create the revised story, with $90.3\%$ average token overlap between them. Similarly, the inferred state and its counterfactual on average show high lexical similarity with $72\%$ token overlap, and both having similar token length.
Additional statistics can be seen in Table~\ref{Table:dataset}.

\begin{table}[t!]
\footnotesize
\centering
\begin{tabular}{m{0.73\linewidth} | m{0.17\linewidth}}
\Xhline{4\arrayrulewidth}
\hline
{\# of unique stories} &  $5,028$ \\
\hline
{Avg. \# of tokens in an inferred state} &  $5.7$ tokens  \\
\hline
{Avg. \# of tokens in a perturbed state} &  $6$ tokens  \\
\hline
{Avg.\# of justification sentences for a state} &  $1.5$  \\
\hline
{Avg. \# of sentences revised in a story} &  $1.48$  \\
\hline
{\% of justification sentences that are revised} &  $90.54\%$  \\
\hline
{\% of revised sentences that were justification} &  $91.9\%$  \\
\hline
{\% tokens in inferred state, common in perturbed state} &  $71.9\%$  
\\ \hline
{\% story tokens common in revised story} &  $90.3\%$  
\\ \Xhline{4\arrayrulewidth}
\end{tabular}
\caption{\small{\dataset{} Dataset Statistics}}
\label{Table:dataset}
\end{table}

\begin{figure}[t!]
    \includegraphics[width=.48\textwidth]{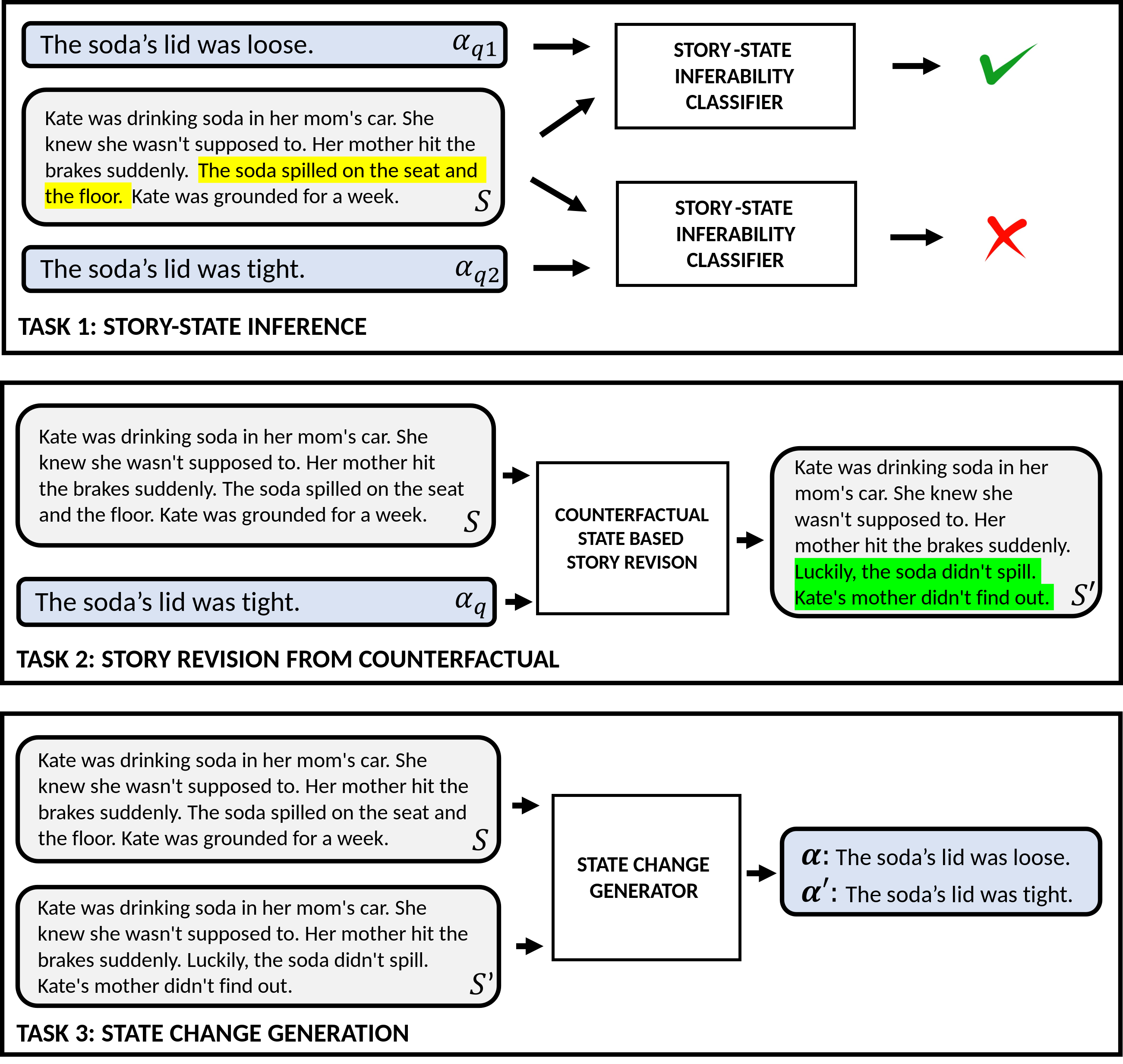}
    \caption{\small{Examples of three PASTA tasks. Input for each is on the left. The boxes indicate systems required to solve the tasks with example output on the right.}}
    \label{Fig:stae_tasks}
\end{figure}
\section{State-based Reasoning Tasks} 
Inferring each component of a \dataset{} 4-tuple requires a different commonsense reasoning ability about a participant's state in a narrative, which enables us to use \dataset{} to test models for these abilities.
As illustrated in Figure~\ref{Fig:stae_tasks}, we introduce three \dataset{} tasks, one classification and two generative, each of which can be used to evaluate current NLP models for the capabilities required to understand a participant's state in a narrative text.
In the subsections below we provide the motivation and formal task definition for each task.
\subsection{\mcq{}}\label{subsec:state_inference} 
We propose a classification task to evaluate a model's ability to understand what state is likely or unlikely to be inferred from a story. 
We deem a state is \emph{likely to be inferred} from a story if a typical human reading the story would conclude that the state is most likely true.
To test this capability in models, we pose the \mcq{} classification task.
\paragraph{Task definition:} Given a story $S$, a `query' state $\alpha_q$,
and a supporting set $\mathbf{s}$, which is a subset of the sentences in $S$, the task is to predict whether $\alpha_q$ is likely to be inferred from $\mathbf{s}$ in the context of $S$.
\paragraph{Effects of data collection on performance:} We provide additional dataset analysis in subsection~\ref{subsec:story_state_inf_eval} to analyze the robustness of our data collection procedure that helped avoid unintended artifacts in the data for this task. 
%
\subsection{\storyrev{}}
A model that can understand participant states in narrative text should also be able to reason about counterfactual states and their potential effects on the narrative. We introduce the \storyrev{} task to address this.
%
\paragraph{Task definition:} Given a story $S$, a participant state $\alpha_q$ that is counterfactual to $S$ (a state that is not consistent with $S$), make minimal revisions to $S$ to generate $S'$ such that $\alpha_q$ is unstated in $S'$ and it can be inferred from $S'$, i.e. $P(\alpha_q|S') \approx 1$ and $P(\alpha_q|S') \gg P(\alpha_q|S)$.
%
%
%
\subsection{\statechange{}}\label{subsec:state_change}
A corollary of being able to reason about the effects of a counterfactual state on the discourse of a narrative is the ability to identify the state changes (and how it changed) which led to the new narrative. In other words, when given a revised story with its original, what original state and its counterfactual explains the change? To assess this, we introduce the \statechange{} task.
\paragraph{Task definition:} Given a story $S$ and its revision $S'$, the task is to generate participant states $\alpha, \alpha'$ that describe the change of state from $S$ to $S'$, i.e. $P(\alpha | S) \gg P(\alpha | S')$ and $P(\alpha' | S') \gg P(\alpha' | S)$.
%
\subsection{Task-specific data creation} 
\label{subsec:task_data_creation}
The three tasks above use the \dataset{} 4-tuple $(S, \alpha, \alpha', S')$ to create task specific data instances in the following manner:\\

\noindent
\textbf{1. \mcq{}}: Let $S=(s_1, \cdots, s_5)$ and $S' = (s'_1, \cdots, s'_5)$.
We create four data instances for the task, 
positive data instances $((S, \mathbf{s}, \alpha), 1)$ and $((S', \mathbf{s'}, \alpha'), 1)$, and 
negative instances $((S, \mathbf{s}, \alpha'), 0)$ and $((S', \mathbf{s'}, \alpha), 0)$.
The supporting set $\mathbf{s}$ for $S$ is $J^S_{\alpha}$, i.e. the minimal set of sentences used to infer $\alpha$ from $S$. 
For $S'$, $\mathbf{s'} = \{s \in \{s'_1, \cdots, s'_5\} | s'_i \neq s_i, \: \forall \: i \in 1 \: \text{to} \: 5\}$ i.e., the set of sentences in $S$ that were changed when revising $S$ to $S'$.

\noindent
\textbf{2. \storyrev{}}: We created two data instances for the task of the form $((S, \alpha'), S')$ and $((S', \alpha), S)$. 

\noindent
\textbf{3. \statechange{}}: We created two data instances for the task of the form $((S, S'), (\alpha, \alpha'))$ and $((S', S), (\alpha', \alpha))$.
{\renewcommand{\arraystretch}{1.1}
\begin{table}
    \footnotesize
\begin{subtable}{0.5\textwidth}
\sisetup{table-format=-1.2}   
\centering
        \begin{tabular}
        { m{0.42\linewidth} m{0.08\linewidth}  |m{0.09\linewidth} |m{0.08\linewidth} } 
        \cline{2-4}
        & \multicolumn{3}{c}{\textbf{\# of examples in prompt}} \\ \hline
        \multicolumn{1}{l|}{\textbf{APPROACH}} & \textbf{5} & \textbf{10} & \textbf{15} \\ \hline \hline
        \multicolumn{1}{l|}{EXPERT CURATED} & \footnotesize{\makecell[cc]{$81.6$}} & \footnotesize{\makecell[cc]{$81.6$}} & \footnotesize{\makecell[cc]{$82.5$}} \\  \hline
        \multicolumn{1}{l|}{RANDOM SELECTION} & \makecell[cc]{$81.2$} & \makecell[cc]{$81.8$} & \makecell[cc]{$82.3$} \\  \hline
        \multicolumn{1}{l|}{NEAREST NEIGHBOR} & \makecell[cc]{$81.7$} & \makecell[cc]{{{${83.3}$}}} & \makecell[cc]{$82.0$} \\ \Xhline{3\arrayrulewidth}
        \end{tabular}
   \caption{\storyrev{}}\label{tab:gpt3_hyperparam1}
\end{subtable}
\bigskip

\begin{subtable}{0.5\textwidth}
\sisetup{table-format=4.0} 
\centering
        \begin{tabular}
        { m{0.42\linewidth} m{0.09\linewidth}  |m{0.08\linewidth} |m{0.08\linewidth} } 
        \cline{2-4}
        & \multicolumn{3}{c}{\textbf{\# of examples in prompt}}
        \\ \hline
        \multicolumn{1}{l|}{\textbf{APPROACH}} & \textbf{5} & \textbf{10} & \textbf{15} \\ \hline \hline
        \multicolumn{1}{l|}{EXPERT CURATED} & \footnotesize{\makecell[cc]{{{${53.0}$}}}} & \footnotesize{\makecell[cc]{$52.8$}} & \footnotesize{\makecell[cc]{$51.9$}} \\  \hline
        \multicolumn{1}{l|}{RANDOM SELECTION} & \makecell[cc]{$51.4$} & \makecell[cc]{$51.2$} & \makecell[cc]{$52.1$} \\  \hline
        \multicolumn{1}{l|}{NEAREST NEIGHBOR} & \makecell[cc]{$52.1$} & \makecell[cc]{$51.5$} & \makecell[cc]{$50.2$} \\ \Xhline{3\arrayrulewidth}
        \end{tabular}
   \caption{\statechange{}}\label{tab:gpt3_hyperparam2}
\end{subtable}
\caption{\small{\textbf{{GPT3} hyperparameter selection.} Few-shot performance (BERTscore) of GPT-3 for combinations of (i) prompt example selection approach (ii) \# of examples in prompt.}
} \label{Table:gpt3_hyperparam}
\end{table}}
%
\begin{figure}[!ht]
    \centering
\scriptsize
\begin{tabular}{p{7.5cm}}
\Xhline{4\arrayrulewidth} 
Task 1. \textbf{\mcq{}} 
\\ \Xhline{4\arrayrulewidth} 
\textbf{Input:} \textbf{infer\_state} \textbf{story:} I live in the north. We usually have snow here for Christmas. This year is different. \hl{* It's 50 degrees out!} \hl{* We will have a green holiday this year.} 
\newline \textbf{state:} I live in the North Pole.</s> 
\\ \Xhline{1\arrayrulewidth}
\textbf{Target Output:} false</s> \\ 
\Xhline{4\arrayrulewidth}
Task 2. \textbf{\storyrev{}}  \\ \Xhline{4\arrayrulewidth} 
\textbf{Input:} \textbf{revise} \textbf{story} 
<extra\_id\_1>: Connor had a very busy workload that day.
<extra\_id\_2>: He forgot to eat breakfast and grabbed a dozen doughnuts on the way.
<extra\_id\_3>: People were greeting him and taking one doughnut each time.
<extra\_id\_4>: When he arrived at his desk, there were no doughnuts left.
<extra\_id\_5>: He went on with his busy day on an empty stomach. \newline 
\textbf{state}: Connor only has six coworkers.</s> 
\\ \Xhline{1\arrayrulewidth}
\textbf{Target Output:} 
<extra\_id\_4>: When he arrived at his desk, half the donuts were gone.
<extra\_id\_5>: He ate a donut before starting work.</s> \\
\Xhline{4\arrayrulewidth}
Task 3. \textbf{\statechange{}}  \\ 
\Xhline{4\arrayrulewidth} 
\textbf{Input:} \textbf{change} \textbf{story1:} Joe was stopping at a drive thru for breakfast.
He ordered a cup of coffee and breakfast sandwich.
Joe was trying to eat and drive at the same time.
Joe accidently dropped his coffee in his lap.
Joe had to go home and change his pants. \newline
\textbf{story2:} Joe was stopping at a drive thru for breakfast.
He ordered a cup of coffee and breakfast sandwich.
Joe was trying to eat and drive at the same time.
Joe accidently dropped his coffee in his lap.
Joe had to go to the hospital.
</s>
\\ \Xhline{1\arrayrulewidth}
\textbf{Target Output:} \textbf{state1:} The coffee wasn't hot enough to seriously burn Joe. \textbf{state2:} The coffee was hot enough to seriously burn Joe.</s>
\\\Xhline{4\arrayrulewidth}
\end{tabular}
\caption{\small{Examples of T5 formats. An asterisk $*$ is prepended to the \hl{supporting set sentences} in the stories in task (1). The output in task (2) includes special tokens like <extra\_id\_4> and  <extra\_id\_5> to indicate revisions to the $4^{th}$ and $5^{th}$ sentences.}}
\label{Table:T5_ip_op}
\end{figure}
\section{Experimental Setup}\label{sec:experiment}
To establish modern baselines and measure their performance, we built benchmark models from {GPT3, T5, BERT}, and {RoBERTa}. This section describes how each was setup for the three tasks.
\subsection{GPT3}
We benchmarked {GPT3} with few-shot prompting \cite{brown2020language} on the two generation tasks (Story Revision and State Change). 
We created prompts with task examples from the training set, followed by an incomplete prompt from the eval set that the model must complete.
For the \storyrev{} task, the prompt included $n$ examples followed by the final query: $(S_1, \alpha'_1, S'_1)\cdots(S_n, \alpha'_n, S'_n)(S_q, \alpha'_q, -)$ where $(S_i, \alpha'_i, S'_i)$ is the $i^{th}$ task example. The model must generate $S'_q$ for the final query ($S_q$,$\alpha'_q, -)$.
Similarly, the \statechange{} task uses a similar prompt: $(S_1, S'_1, \alpha_1, \alpha'_1)\cdots(S_n, S'_n, \alpha_n, \alpha'_n)(S_q, S'_q, -, -)$. 
\par
To select prompt examples, we tried three approaches. \textbf{(i) EXPERT CURATED:} We selected a fixed set of diverse, unambiguous examples that requires multi-step reasoning and covers different type of states, and used the same prompt examples for all the query instances, \textbf{(ii) RANDOM SELECTION:} We randomly selected examples, \textbf{(iii) NEAREST NEIGHBOR}~\cite{liu-etal-2022-makes}: For each query instance, we selected examples that were most similar to it. For this, we computed the cosine similarity between the \texttt{[CLS]} representation of the instances obtained from {RoBERTa-large} fine-tuned on the \mcq{} task. For each approach, we tried creating prompts with $5\text{, } 10$ and $15$ examples.
Prompt examples were selected from a set of 200 high-quality, expert-selected instances drawn from the training set, similar to~\citet{West2022SymbolicKD}.
%

We treat the number of prompt examples and their selection as hyperparameter combinations, and evaluated each of them on 200 random samples from the validation set. Since human evaluations are expensive, we use BERTScore, which has the highest correlation with human evaluated validity of output, among the automatic metrics we tried (see Table~\ref{Table:pasta_corr}). Table~\ref{Table:gpt3_hyperparam} shows that the combinations perform roughly similar but there is a two point gap between the best and the worst combination. For the \storyrev{} task, we use NEAREST NEIGHBOR with $10$ prompt examples, and for the ~\statechange{} task, we use EXPERT CURATED with $5$ examples.


%
We used the \texttt{text-davinci-002} {GPT3} model for both tasks. We set the generation temperature parameter to $0.9$, frequency penalty to $0.5$, and maximum generation length to $100$.
%
\subsection{T5}  
We benchmarked {base} ({T5-b}) and {large} ({T5-l}) variants of T5 on all three state-based tasks by fine-tuning them on the task-specific instances created from \dataset{}, as explained in Section~\ref{subsec:task_data_creation}. 
Examples of T5 input-output format for each task are shown in Figure~\ref{Table:T5_ip_op}. 
For all the tasks, {T5-b} and {T5-l} were trained for $7$ and $5$ epochs, respectively. 
For model training, we used AdamW \cite{loshchilov2017decoupled} optimizer with a learning rate of $10^{-4}$ and weight decay of $10^{-6}$. For {T5-l} the batch-size for tasks 1, 2 and 3 were $8$, $4$ and $4$ respectively. Whereas for {T5-b}, the corresponding batch-sizes were $16$, $12$, and $10$. For text generation, we used nucleus sampling with $0.93$ top-p; $100$ as max generation length.
%
%
\subsection{BERT~\textbackslash~RoBERTa}
We benchmarked {base} ({BERT-b}) and {large}  ({BERT-l}) variants of {BERT-uncased}, and {base}  ({RoBERTa-b}) and {large} ({RoBERTa-l}) of {RoBERTa} on only the \mcq{} task since they are non-generative models.
The input format for the models is identical to that of {T5}. For all the models, we used AdamW optimizer with a learning rate of $5\text{e}{-6}$ and weight decay of $1\text{e}{-6}$. 
The large and base models were trained for $5$ and $7$ epochs respectively.

{T5-b, BERT-b, BERT-l, RoBERTa-b} and {RoBERTa-l} were trained on an NVIDIA-TITAN-X 24GB, and {T5-l} was trained on an NVIDIA-A6000 48GB GPU.

\section{Results and analysis}
We now analyze the performance of these recent language models on the three \dataset{} tasks.
{\renewcommand{\arraystretch}{1.2}
\begin{table}[!ht]
    \small
    \begin{center}
        \begin{tabular}
        { m{0.28\linewidth}| m{0.25\linewidth}  |m{0.25\linewidth}  } 
        \Xhline{3\arrayrulewidth}
         & \textbf{\footnotesize{Accuracy (\%)}} & \textbf{\footnotesize{Contrastive Accuracy (\%)}}  \\ \hline \hline
        \footnotesize{\makecell[l]{\textbf{BERT-b}}} & \makecell[cc]{$73.8 \pm 0.3$} & \makecell[cc]{$64.0 \pm 0.5$} \\  \hline        
        \footnotesize{\makecell[l]{\textbf{T5-b}}} & \makecell[cc]{$79.8 \pm 0.6$} & \makecell[cc]{$70.7\pm 0.5$} \\  \hline
        \footnotesize{\makecell[l]{\textbf{RoBERTa-b}}} & \makecell[cc]{$81.2 \pm 0.6$} & \makecell[cc]{$73.0 \pm 0.8$} \\  \Xhline{3\arrayrulewidth}
        \footnotesize{\makecell[l]{\textbf{BERT-l}}} & \makecell[cc]{$77.5 \pm 0.4$} & \makecell[cc]{$68.7 \pm 0.7$} \\  \hline
        \footnotesize{\makecell[l]{\textbf{T5-l}}} & \makecell[cc]{$83.1 \pm 0.9$} & \makecell[cc]{$75.3\pm 1.4$} \\  \hline
        \footnotesize{\makecell[l]{\textbf{RoBERTa-l}}} & \makecell[cc]{$89.1 \pm 0.4$} & \makecell[cc]{$83.7 \pm 0.5$} \\  \Xhline{3\arrayrulewidth}
        \footnotesize{\hyperref[par:human_eval]{\textbf{\footnotesize{Human$^\bigstar$}}}} & \makecell[cc]{$96.9$} & \makecell[cc]{$94.2$} \\  \Xhline{3\arrayrulewidth}
        \end{tabular}
    \end{center}
  \caption{\small{\textbf{\mcq{} - Model evaluation}: Accuracy is the \% of instances where the model made correct predictions. Contrastive Accuracy gives a credit to the model if it correctly predicts the inferability for both the inferred and counterfactual states for a story.
  }
  }
  \label{Table:state_inference_main_results}
\end{table}}
\subsection{\mcq{} }\label{subsec:story_state_inf_eval}
We evaluated model performance with standard accuracy and contrastive accuracy. In contrastive accuracy, the model gets a point only if it makes correct predictions for both inferred and the counterfactual states for a story. For all models, we train with five random seeds and report their average performance with standard deviation.
\paragraph{Human evaluation:}\label{par:human_eval}
We conducted human evaluation on the task instances (see Section~\ref{subsec:task_data_creation}.1) created from \dataset{}.
We randomly selected $200$ 4-tuples from the test set\footnote{Model performance for this test subset differed by $<0.5\%$ from that of the overall test set}, and created $800$ story-state inference instances from them.
Each task instance, $(S, \alpha_q, \mathbf{s})$ was evaluated by three crowd workers who rated the likelihood of inferring $\alpha_q$ from $\mathbf{s}$ in context of $S$, on a 5-point Likert scale - {Extremely unlikely, Unlikely, Cannot Say, Likely, and Extremely likely}.
We threshold the Likert value to a binary 0/1 value with the mapping, \{Extremely unlikely to Cannot Say\} $\rightarrow$ 0 and rest $\rightarrow$ 1.
The human prediction for an instance was computed by majority voting, which along with its true label was used to compute the human performance.\footnote{The instance label assignment is explained in Section~\ref{subsec:task_data_creation}} %
\paragraph{\mcq{} is a hard task}
Table~\ref{Table:state_inference_main_results} shows that even for just standard accuracy, there is room for improvement ($7.8\%$) when comparing the best performing model (RoBERTa-l) to humans on this simple binary classification task.
This performance gap further increases to $10.5\%$ when considering contrastive accuracy.
Increasing model size from base to large yields $3.7\% \text{ (BERT)} \text{ to } 8\% \text{ (RoBERTa)}$ gains on standard accuracy.
For contrastive measure, both base and large variants of each models fare substantially worse, with performance drops ranging from $5.4\%$ (RoBERTa-l) to $9.8\%$ (BERT-b). For humans, the corresponding performance drop is only $\sim2.7 \%$.
This suggests that predicting whether a state is likely to be inferred from a story is difficult for these LLMs, even when fine-tuning on a relatively large number of examples.
%
{\renewcommand{\arraystretch}{1.2}
\begin{table}[!ht]
    \small
    \begin{center}
        \begin{tabular}
        { m{0.25\linewidth}| m{0.25\linewidth}  |m{0.25\linewidth}  } 
        \Xhline{3\arrayrulewidth}
         & \textbf{\footnotesize{Accuracy (\%)}} & \textbf{\footnotesize{Contrastive Accuracy (\%)}}  \\ \hline \hline
        \footnotesize{\makecell[l]{\textbf{BERT-l}}} & \makecell[cc]{$74.9 \pm 0.3$} & \makecell[cc]{$64.6 \pm 0.3$} \\  \hline
        \footnotesize{\makecell[l]{\textbf{T5-l}}} & \makecell[cc]{$79.6 \pm 0.6$} & \makecell[cc]{$69.8\pm 1.0$} \\  \hline
        \footnotesize{\makecell[l]{\textbf{RoBERTa-l}}} & \makecell[cc]{$86.7 \pm 0.4$} & \makecell[cc]{$80.4\pm 0.6$} \\  \hline
        \footnotesize{\hyperref[par:human_eval]{\textbf{\footnotesize{Human$^\bigstar$}}}} & \makecell[cc]{$93.5$} & \makecell[cc]{$88.9$} \\  \Xhline{3\arrayrulewidth}
        \end{tabular}
    \end{center}
  \caption{\small{\textbf{\mcq{} - without the justification sentences}: Model performance on a harder variant of the \mcq{} task, where they don't have direct access to the justification sentences of a state when predicting it's inferability for a story.}
  }
  \label{Table:state_inference_wo_support_main_results}
\end{table}}
\par
%
We also analyze the performance of the models when they don't have direct access to the justification sentence information in the story. We fine-tuned large variants of the three baseline models on this task.
From Table~\ref{Table:state_inference_main_results} to \ref{Table:state_inference_wo_support_main_results}, we see that the task performance drops across all the models on both evaluation metrics, with a $2.4\%$ to $3.5\%$ drop in accuracy, and $3.3\%$ to $5.5\%$ in contrastive accuracy. The gap between the human performance and the best performing model is still substantial.
This shows that justification sentences are indeed important to solve the task, but the models often still make reasonable decisions without them.
{\renewcommand{\arraystretch}{1.2}
\begin{table}[t]
    \small
    \begin{center}
        \begin{tabular}{ m{0.18\linewidth} m{0.19\linewidth}  | m{0.19\linewidth} | m{0.2\linewidth}  } 
        \cline{2-4}\cline{2-4}
        & \multicolumn{3}{c}{\textbf{Test data}} \\ \Xhline{3\arrayrulewidth}
        \multicolumn{1}{c|}{\textbf{Train data}} &
         \makecell[cc]{$\mathcal{D}^{te}_1$} & \makecell[cc]{$\mathcal{D}^{te}_2$} & \makecell[cc]{$\mathcal{D}^{te}_1 \cup \mathcal{D}^{te}_2$}  \\ \hline \hline
         \multicolumn{1}{c|}{$\mathcal{D}^{tr}_1$} & $90.2 (84.4)$ & $79.1 (70.8)$ & $84.8 (77.9)$ \\ \hline
         \multicolumn{1}{c|}{$\mathcal{D}^{tr}_2$} & $81.5 (73.6)$ & $88.1 (82.6)$ & $84.8 (78.1)$ \\ \hline
     \multicolumn{1}{c|}{$\mathcal{D}^{tr}_1 \cup \mathcal{D}^{tr}_2$} &$90.2 (85.3)$ & $88.1 (82.3)$ & $89.1 (83.7)$ \\ \Xhline{3\arrayrulewidth}
        \end{tabular}
    \end{center}
  \caption{\small{\textbf{\mcq{} - Dataset Analysis:} 
  Accuracies and contrastive accuracies of {RoBERTa-large} when trained and tested on dataset partitions created from original stories ($\mathcal{D}_1$), modified stories ($\mathcal{D}_2$), and their union ($\mathcal{D}_1\cup \mathcal{D}_2$). $tr \text{ and } te$ denotes the corresponding training and test splits.}
  }
  \label{Table:state_inference_artifiact}
\end{table}}

\paragraph{Importance of the data collection design:}
It is important to note that we included contrastive examples in our train-set. To illustrate its importance,  we trained a model on the dataset created from just the original stories ($\mathcal{D}_1$ = $\{((S_i, \mathbf{s_{i}}, \alpha_i), 1), ((S_i, \mathbf{s_{i}}, \alpha_i'), 0) \}^N_{i=1}$), and another on the modified stories ($\mathcal{D}_2 = \{((S'_i, \mathbf{s'_{i}}, \alpha_i'), 1), ((S'_i, \mathbf{s'_{i}}, \alpha_i), 0) \}^N_{i=1}$).
We then trained and tested on these different dataset, results of which are reported in Table~\ref{Table:state_inference_artifiact}.

Generalization accuracy is significantly worse if we had only constructed positive and negative states for a collection of stories. For example, training on $\mathcal{D}_1^{tr}$ and testing on $\mathcal{D}_2^{te}$ leads to an $11.1\%$ drop in accuracy compared to in-distribution test on $\mathcal{D}_1^{te}$. For $\mathcal{D}_2^{tr}$, the corresponding drop is $6.6\%$, which supports the quality of our stories/states and shows that both original and counterfactual state inferences are learnable. Had we not collected the revised story, then models could potentially learn artifact-based heuristics (e.g. guessing whether the state is original or modified) resulting in the lack of generalization that we observe here. Because \dataset{} includes the revised stories, we can train on the full dataset $\mathcal{D}^{tr} = D_1^{tr} \cup D_2^{tr}$, and see that the performance is uniform across the different test partitions. This highlights the challenges in constructing negative examples for such tasks and the importance of including contrastive examples for both training and test for proper generalization.
\begin{table}[!h]
    \small
    \begin{center}
        \begin{tabular}{  m{0.19\linewidth} | m{0.145\linewidth}| m{0.12\linewidth} | m{0.14\linewidth} | m{0.12\linewidth}  } 
        \Xhline{3\arrayrulewidth}
    & \multicolumn{3}{c|}{\textbf{\footnotesize{Acceptability}}} & \\ \cline{2-4}
    & 
    \textbf{\footnotesize{\makecell[cc]{\% \\ Inferable \\(A)}}} & 
    \textbf{\footnotesize{\makecell[cc]{\% \\ Logical \\(B)}}} & 
    \textbf{\footnotesize{\makecell[cc]{\% ALL \\(A \& B)}}} & \textbf{\footnotesize{\makecell[cc]{Minimal \\ Revision}}} \\ 
    \hline \hline
        \footnotesize{\textbf{GPT3 - FS}} &  $50$ & $86$ & $48.5$  & $86.33$  \\
        \hline
        \footnotesize{\textbf{T5-b FT}} &  $41.0$ & $77.0$ & $34.0$ & $91.39$ \\ 
        \cline{1-5}
        \footnotesize{\textbf{T5-l FT}} &  $58.5$ & $84.0$ & $54.0$ & $89.17$ \\ 
        \Xhline{3\arrayrulewidth}
        \end{tabular}
    \end{center}
  \caption{\small{\textbf{\storyrev{} - Human evaluation}: $\%$ of model generated stories that satisfy evaluation criteria. \textbf{Inferable} and \textbf{Logical} assure coherence of the revisions with the required entity state. \textbf{ALL} indicates generations that satisfied both the criteria. \textbf{FS} means few-shot learning, \textbf{FT} means finetuned on \dataset{} dataset.}
  }
  \label{Table:story_rev_human_eval}
\end{table}
\subsection{\storyrev{}}\label{subsec:story_rev_result}
This generation task requires the model to revise a given story, such that the revised story is consistent with the given counterfactual participant state.
We use human judgments to evaluate the revised stories because reference-based automatic evaluation metrics (e.g. BLEU~\cite{papineni2002bleu}, BERTscore~\cite{zhang2019bertscore} etc.) are inadequate for multiple reasons: 
(i) valid revised stories often exist that are different from the references, 
(ii) original and revised stories overlap heavily which can skew the metrics, and
(iii) small lexical changes that don't change automatic metrics can affect logical consistency. 
We thus evaluate generation quality using our proficient workers from Section~\ref{subsec:quality_control}. 

We compare performance of the models on a subset of $200$ test instances chosen at random. We evaluated them for quality on three metrics:
(1) \textbf{Inferable}: how likely is it for the given state $\alpha'$ to be true at any point in the revised story $S'$? This was rated on a 5-point Likert scale, which we thresholded to a 0/1 value (1 means inferrable).
(2) \textbf{Logical}: is the generated story $S'$ logically correct? This was a \texttt{YES/NO} question.  
(3) \textbf{Minimal revison}: what is the degree of revision made to $S$ to generate $S'$? This was rated on a 5-point Likert-scale, with 4 indicating minimal revision and 0 an entirely new story.
Higher scores indicate higher similarity between $S$ and $S'$. %
\textbf{Inferability} and \textbf{Logical} decide the ultimate correctness of a response. 
We calculated an overall model acceptability score (\textbf{ALL} in Table~\ref{Table:story_rev_human_eval}) by finding the percentage of model output that were both logical, and the input state can be inferred from them.

Table~\ref{Table:story_rev_human_eval} shows that T5-l outperforms T5-b and GPT3 on the acceptability (\textbf{ALL}) of generated outputs by a large margin of $20\%$ and $5.5\%$, respectively. 
GPT3 has the best performance on logical validity of the generated output with T5-l lagging behind by only $2\%$, but only $50\%$ of GPT3's output satisfy the inferability criteria.
In fact  all the models have low inferability score, which brings down their overall acceptability score.
T5-b has the best performance on the `minimal revision' made to the original story, however this was not a primary metric of concern and there is always a trade-off between doing well on this score and generating an acceptable result. For example, revising a story conditioned on a counterfactual that is connected to entities in a different part of the story might require substantial revisions.


Overall, only $54\%$ of the output generated by the best model, T5-l, are  acceptable, indicating that the task is challenging and there is large room for improvement. Our results with GPT3 were based on few-shot prompting where we treated its design choices as a modelling hyperparameter that were chosen based on automatic metric performance on the validation set. 
Few-shot performance of GPT3-scale models depends heavily on prompt engineering, so this direction may require further investigation.
\begin{table}[!htp]
    \small
    \begin{center}
        \begin{tabular}{  m{.165\linewidth}| m{0.145\linewidth}| m{0.15\linewidth}| m{0.135\linewidth}| m{0.135\linewidth} } 
        \Xhline{3\arrayrulewidth}
        \textbf{{\makecell[lc]{Model}}} 
        &  \textbf{\footnotesize{\makecell[c]{\% Valid \\ Attribute \\(A)}}}
        & \textbf{\footnotesize{\makecell[c]{\% Valid \\ Inferable \\(B)}}}
        & \textbf{\footnotesize{\makecell[c]{\% Not \\ in Story \\(C)}}}
        & \textbf{\footnotesize{\makecell[c]{\% ALL \\A, B \& C}}}\\ \hline \hline
        \footnotesize{\textbf{GPT3 FS}} &  $86.5$ & $67.46$ & $81.5$  & $47.7$\\
        \hline
        \footnotesize{\textbf{T5-b FT}} &  $96.75$ & $41.0$ & $90.25$ & $35.17$\\ \cline{1-5}
        \footnotesize{\textbf{T5-l FT}} &  $99.25$ & $58.75$ & $97.0$ & $55.50$\\ 
        \Xhline{3\arrayrulewidth}
        \end{tabular}
    \end{center}
  \label{Table:state_change_human_eval}
  \caption{\small{\textbf{\statechange{} - Human evaluation}:
   \% of output that satisfy the evaluation criteria. \textbf{Valid attributes} ensures that the generated states describe entity attributes, not actions. \textbf{Valid inferability} indicates if both states can be inferred from their correct stories. \textbf{Not in Story} assures that the states are unstated in the stories. \textbf{ALL} indicates the \% of generations that satisfy all the these criteria.}
  }
  \label{Table:state_change_results}
\end{table}
\subsection{\statechange{}}
In this task, for given stories $S$ and $S'$, the model generates the two states $\alpha$ and $\alpha'$.
As in the previous task, we do a human evaluation of a randomly selected set of $200$ model generated outputs.

Model outputs were evaluated on the following metrics: (1) \textbf{Valid Attribute:} do the generated states $\alpha$, $\alpha'$ describe entity attributes? This was a \texttt{YES} or \texttt{NO} question. (2) \textbf{Valid Inferability:} are generated states $\alpha$ and $\alpha'$ inferable from $S$ and $S'$, but not from $S'$ and $S$, respectively? Workers rated $\alpha$ and $\alpha'$'s likelihood of being inferred independently, on a 5-point Likert scale, which was thresholded to a 0/1 value. For instance, if $\alpha$ is inferred from $S$, then $L_{S\alpha}=1$ (otherwise 0). Based on these scores, the inferability change for $\alpha$ is computed (1 for valid, 0 for invalid) using $max(0, L_{S\alpha} - L_{S'\alpha})$. (3) \textbf{Not in Story:} are $\alpha$ and $\alpha'$ unstated in both $S$ and $S'$? This was a multiple-choice question with 4 choices, 3 corresponding to the state being present in either one or both the stories, and the 4th for neither of the stories. An output $(\alpha, \alpha')$ gets full credit on a metric if both $\alpha$ and $\alpha'$ are correct for that metric, half if only one of them ($\alpha$ or $\alpha'$) is correct, and 0 otherwise. \textbf{ALL} indicates full credit on all three metrics.

Table~\ref{Table:state_change_results} shows the results. T5-l in general outperforms both T5-b and GPT3 on all the metrics except the \textbf{Valid Inferability}, where GPT3 outperforms the other models by a large margin
. Interestingly, GPT3 is the worst performing model on \textbf{Valid Attribute} and \textbf{Not in Story}. 
This indicates that GPT3 is loosely "cheating" by copying text in the story itself, which of course is inferable, but violates the task's requirement of an implicit state.
Overall, the best acceptability score (\textbf{ALL} in Table~\ref{Table:state_change_results}) is only $55.5\%$, which suggests that generating an output that satisfies all the criteria for a quality state change is an interesting challenge.
\subsection{Automatic Evaluation for Generative Tasks}\label{sec:auto_eval}
For the two generative tasks, we reported human evaluation results for the best analysis (prior sections).
However, since human evaluation is expensive, we include here the results from three automatic metrics: {GLEU} \cite{Wu2016GooglesNM}, {ROUGE} \cite{lin-2004-rouge}, and {BERTscore} \cite{zhang2019bertscore}. 
For {GLEU}, we consider 1 to 4-grams overlap between the output and reference. 
We report {ROUGELsum} for the \storyrev{} task since it is computed over the entire story, and the sentence level {ROUGEL} metric for \statechange{}. 
From Table~\ref{tab:story_rev_eval} and \ref{tab:st_change_eval}, we can observe that even for automatic metrics, {T5-l} is still the best performing model on both tasks. 

\begin{table}
    \small
\begin{subtable}{0.5\textwidth}
\sisetup{table-format=-1.2}   
\centering
        \begin{tabular}
        { m{0.17\linewidth} | m{0.22\linewidth} |m{0.165\linewidth} |m{0.195\linewidth} } 
        \Xhline{3\arrayrulewidth}
         & \footnotesize{\makecell[cc]{\textbf{BERTscore}}} & \footnotesize{\makecell[c]{\textbf{GLEU}}} & 
         \footnotesize{\makecell[c]{\textbf{rougeLsum}}} \\ \hline \hline
        \footnotesize{\makecell[l]{\textbf{GPT3 FS}}} & 
        \makecell[cc]{$80.7$} & \makecell[cc]{$69.7$} & \makecell[cc]{$79.6$} \\  \hline
        \footnotesize{\makecell[l]{\textbf{T5-b FT}}} & \makecell[cc]{$81.6$} & \makecell[cc]{$73.2$} & \makecell[cc]{$81.7$} \\  \hline
        \footnotesize{\makecell[l]{\textbf{T5-l FT}}} & \makecell[cc]{$82.1$} & \makecell[cc]{$73.5$} & \makecell[cc]{$81.7$} \\ \Xhline{3\arrayrulewidth}
        \end{tabular}
   \caption{\storyrev{}}\label{tab:story_rev_eval}
\end{subtable}

\bigskip
\begin{subtable}{0.5\textwidth}
\sisetup{table-format=5.0} 
\centering
        \begin{tabular}
        { m{0.17\linewidth} | m{0.22\linewidth} |m{0.17\linewidth} |m{0.19\linewidth} } 
        \Xhline{3\arrayrulewidth}
         & \footnotesize{\makecell[cc]{\textbf{BERTscore}}} & \footnotesize{\makecell[c]{\textbf{GLEU}}} & 
         \footnotesize{\makecell[c]{\textbf{ROUGEL}}} \\ \hline \hline
        \footnotesize{\makecell[l]{\textbf{GPT3 FS}}} & {\makecell[cc]{$55.4$}} & 
        {\makecell[cc]{$11.6$}} & 
        {\makecell[cc]{$28.9$}} \\  \hline
        \footnotesize{\makecell[l]{\textbf{T5-b FT}}} & \makecell[cc]{$54.4$} & 
        \makecell[cc]{$11.7$} & 
        \makecell[cc]{$29.5$} \\  \hline
        \footnotesize{\makecell[l]{\textbf{T5-l FT}}} & \makecell[cc]{$56.9$} & 
        \makecell[cc]{$13.4$} & 
        \makecell[cc]{$32.4$} \\ \Xhline{3\arrayrulewidth}
        \end{tabular}
   \caption{\statechange{}}\label{tab:st_change_eval}
\end{subtable}
\caption{\small{\textbf{Automatic evaluation for generative tasks: } Model performance on the generative tasks using BERTscore, GLEU and ROUGE based metrics.}} \label{Table:pasta_autmatic_eval}
\end{table}
\begin{table}[!h]
    \footnotesize
    \begin{center}
        \begin{tabular}{m{0.12\linewidth} | m{0.22\linewidth} |m{0.22\linewidth} |m{0.22\linewidth}}
        \Xhline{3\arrayrulewidth}
         & \multicolumn{1}{c|}{\footnotesize{\textbf{BERTscore}}} 
         & \multicolumn{1}{c|}{\footnotesize{\textbf{GLEU}}} 
         & \multicolumn{1}{c}{\footnotesize{\textbf{ROUGE}}} \\ \hline \hline
        \footnotesize{\makecell[l]{\textbf{Task-1}}} &
        \small{\makecell[cc]{$.21$ $($6\text{e-}7$)$}} & 
        \small{\makecell[cc]{$.14$ $($9\text{e-}4$)$}} & 
        \small{\makecell[cc]{$.13$ $($2\text{e-}4$)$}} \\  \hline
        \footnotesize{\makecell[l]{\textbf{Task-2}}} & \small{\makecell[cc]{$.27$ $($4\text{e-}22$)$}} & 
        \small{\makecell[cc]{$.21$ $($1\text{e-}13$)$}} & 
        \small{\makecell[cc]{$.22$ $($1\text{e-}15$)$}} \\ \Xhline{3\arrayrulewidth}
        \end{tabular}
    \end{center}
    \caption[Caption for LOF]{\small{\textbf{Pearson Correlation} between automatic metric score of a model output and its validity as determined by humans. Numbers in parenthesis are p-value\protect\footnotemark~for the null hypothesis that they are uncorrelated. \textbf{Task-1} is \storyrev{}, and \textbf{Task-2} is \statechange{}.}}
  \label{Table:pasta_corr}
\end{table}
\footnotetext{Lower the p-value, higher is the confidence for rejecting the null hypothesis.}
To further analyze automatic metrics as an alternative to human evaluation, we computed the correlation between them. We computed the Pearson correlation between the automatic metric score of an output and its validity as determined by humans. The results are reported in Table~\ref{Table:pasta_corr}.
The numbers in parenthesis are the p-values for the null hypothesis ($95\%$ confidence interval) that they are uncorrelated. We observed that {BERTscore} has the highest correlation with human evaluated validity for both tasks, outperforming other metrics by a substantial margin. The low p-value further indicates that the correlation is statistically significant. However, since the correlation is low, we strongly recommend using human evaluations, and only use {BERTscore} as an alternative where human evaluation is expensive.
\renewcommand{\arraystretch}{1.1}{
\begin{table}
    \footnotesize
    \centering
    \begin{tabular}{  m{.46\linewidth}| m{0.18\linewidth}| m{0.18\linewidth} } 
    \Xhline{3\arrayrulewidth}
    \multirow{2}{*}{\textbf{{\makecell[c]{Task}}}} & \multicolumn{2}{c}{\textbf{Gwet's coefficient}} \\ \cline{2-3}
    & \makecell[c]{\textbf{Coeff}} & \makecell[c]{\textbf{StdErr}} \\\hline
    
    \mcq{} & \makecell[c]{{0.81}} & \makecell[c]{{0.01}} \\ \hline
    
    \makecell[l]{Story Revision from \\ a Counterfactual} & \makecell[c]{{0.72}} & \makecell[c]{{0.02}} \\ \hline 
    
    \statechange{} & \makecell[c]{{0.76}}& \makecell[c]{{0.01}} \\
    \Xhline{3\arrayrulewidth}
    \end{tabular}
    \caption[Caption for LOF]{\small{\textbf{Inter-Annotator Agreement 
    } for human evaluation for the three tasks. \textbf{Coeff} is the calculated IAA coefficients, and \textbf{StdErr} is the standard error.}}
    \label{Table:IAA}
\end{table}}
\subsection{Inter-Annotator Agreement}
 We measure the inter-annotator-agreement (IAA) for the human workers using Gwet's Agreement Coefficient~\cite{gwet2008computing, gwet2014handbook}, which is a type of generalized Kappa statistic.\footnote{Gwet's normalizes the probability of observed agreement with a percent chance agreement that is the propensity of raters to agree on hard-to-rate instances~\cite{gwet2014handbook}.} 
 Its interpretation is similar to generalized kappa~\cite{viswanathan2012development}, with $0.6-0.8\equiv$ substantial and $\geq0.8\equiv$ almost perfect agreement.
 We use Gwet's coefficient because it is robust to the paradoxical behaviors~\cite{wongpakaran2013comparison, gwet2014handbook} seen in the commonly used IAA Kappa metrics (e.g. Cohen's and Fleiss). This paradoxical behavior of these metrics can lead to their IAA coefficients being lower even when the agreement is strong~\cite{feinstein1990high, byrt1993bias}. 


\paragraph{Crowd workers IAA} Table~\ref{Table:IAA} shows the IAA coefficient for the tasks and their standard errors. For each task, we computed the IAA coefficient for their respective evaluation metrics on their original scale (pre-thresholding\footnote{Note that thresholding was only done for ordinal scale metrics.}), which were then averaged to obtain the overall task scores.
We computed the unweighted IAA coefficient for an evaluation metric if it was nominal, with quadratic weight if it was ordinal.
As can be observed from the table, the crowd worker have strong agreement for both the generative tasks and almost perfect agreement for the classification task.

\paragraph{Experts IAA} 
The two experts in Section~\ref{subsec:quality_control} were responsible for accepting or rejecting a worker response for the \dataset{} creation. 
To measure their IAA, we created a pool of $200$ \dataset{} instances that included both accepted and rejected instances. 
The experts had a Gwet's coefficient of $0.87$ and agreed on $93.5\%$ of those $200$ instances.
\section{Discussion} 

Here we discuss the main challenges and error analyses that highlight areas for future work.

\subsection{Challenges} 
The key challenge common across all tasks is access to diverse types of knowledge (e.g. commonsense, numerical, factual, etc.), as well as the ability to combine and reason with them. For example, task 1 in Figure~\ref{Table:T5_ip_op} requires factual knowledge about the temperatures in the North Pole, commonsense about snow and Christmas, and the ability to combine these when reasoning to detect the incompatibility of the input state.

The Story Revision Task has the added challenge of a model identifying the parts of the input story that are inconsistent with the counterfactual state, and then finally generating logically coherent text. For instance in Figure~\ref{Table:T5_ip_op} task 2, based on the input story and state, the model must first infer from sentences 2-4 that Connor had $12$ coworkers. Then to generate the revised story, it also needs to reason about how the world state gets affected if there were fewer people than the number of doughnuts (e.g., now Connor would have some doughnuts leftover).

The main challenge in the \statechange{} task is that there can be numerous plausible state pairs that are compatible with both stories, but they don't reflect a pertinent state change. Each state needs to be incompatible with one of the stories and compatible with the other, and this differentiation is a big challenge for any model.
For example in Figure~\ref{Table:T5_ip_op} task 3, the observable difference between the stories is the outcome from coffee spilling on Joe. Using abductive reasoning with commonsense knowledge about temperature, one can easily infer that the change in state leading to a different ending comes from the coffee's temperature.
  

%
\subsection{Error Analysis}\label{subsec:error_anl} 

We analyze the model's errors on $200$ randomly selected instances from the validation set.
{\renewcommand{\arraystretch}{1.2}
\begin{table}[!ht]
\footnotesize
    \begin{center}
        \begin{tabular}
        { m{0.19\linewidth} | m{0.28\linewidth}| m{0.14\linewidth}  |m{0.18\linewidth}  } 
        \Xhline{3\arrayrulewidth}
         \multicolumn{1}{c|}{} & \textbf{\makecell[cc]{State Type}} & \textbf{\makecell[cc]{Acc. $\%$}} & \textbf{\makecell[cc]{Contrastive \\ Acc. $\%$}}  \\ \hline \hline
        \multirow{4}{*}{\makecell[l]{\textbf{BERT-l}}} 
        & All - $100\%$ & \makecell[cc]{$79$} & \makecell[cc]{$71.5$} \\  \cline{2-4}
        & Societal - $14.5\%$ & \makecell[cc]{$70.7$} & \makecell[cc]{$62.1$} \\  \cline{2-4}
        & Emotional - $54\%$ & \makecell[cc]{$80.8$} & \makecell[cc]{$74.1$} \\  \cline{2-4}
        & Physical - $31.5\%$ & \makecell[cc]{$79.8$} & \makecell[cc]{$71.4$} \\  \Xhline{3\arrayrulewidth}  
        \multirow{4}{*}{\makecell[l]{\textbf{T5-l}}} 
        & All - $100\%$ & \makecell[cc]{$85.7$} & \makecell[cc]{$80.5$} \\  \cline{2-4}
        & Societal - $14.5\%$ & \makecell[cc]{$81.9$} & \makecell[cc]{$75.9$} \\  \cline{2-4}
        & Emotional - $54\%$ & \makecell[cc]{$87$} & \makecell[cc]{$81.5$} \\  \cline{2-4}
        & Physical - $31.5\%$ & \makecell[cc]{$85.3$} & \makecell[cc]{$81$} \\ \Xhline{3\arrayrulewidth}  
        \multirow{4}{*}{\makecell[l]{\textbf{RoBERTa-l}}} 
        & All - $100\%$ & \makecell[cc]{$90.6$} & \makecell[cc]{$86.6$} \\  \cline{2-4}
        & Societal - $14.5\%$ & \makecell[cc]{$83.6$} & \makecell[cc]{$77.6$} \\  \cline{2-4}
        & Emotional - $54\%$ & \makecell[cc]{$93.5$} & \makecell[cc]{$89.4$} \\  \cline{2-4}
        & Physical - $31.5\%$ & \makecell[cc]{$88.9$} & \makecell[cc]{$86.5$} \\ \Xhline{3\arrayrulewidth}  
     \end{tabular}
    \end{center}
  \caption{\small{\textbf{\mcq{}}: Model performance for predicting the state inferability of different type of states.}
  }
  \label{Table:task8-error}
\end{table}}
\renewcommand{\arraystretch}{1.0}{
\begin{table}
    \footnotesize
    \centering
    \begin{tabular}{l|c}
    \Xhline{3\arrayrulewidth}
    \multicolumn{1}{c|}{\textbf{Error Category}} & \textbf{Percentage} \\
    \hline
        Illogical revised story & 30.1\\
        Irrelevant change & 27.7\\ 
        Contradiction & 20.5\\
        Input state not entailed & 20.5\\
        State explicit in the revision & 1.2\\
    \Xhline{3\arrayrulewidth}
    \end{tabular}
    \caption{\small{Categories of different type of errors for the \textbf{\storyrev{}} task.}}
    \label{Table:task6-error}
\end{table}}
\begin{table}[t]
\footnotesize
\centering
\begin{tabular}{l|c}
\Xhline{3\arrayrulewidth}
\multicolumn{1}{c|}{\textbf{Error Category}} & \textbf{Percentage} \\
\hline
Contradiction & 37.4\\
Irrelevant states & 35.2\\
Illogical state change & 13.2\\
States reversed & 4.4\\
No change in state & 4.4\\
State directly stated in the story & 4.4\\
Actions instead of states & 1.1\\
\Xhline{3\arrayrulewidth}
\end{tabular}
\caption{\small{Categories of different type of errors for the \textbf{\statechange{}} task.}}
\label{Table:task7-error}
\end{table}
\begin{figure}[t]
\scriptsize
\begin{tabular}{p{7.5cm}}
\Xhline{4\arrayrulewidth} 
1. Irrelevant change  
\\ \Xhline{4\arrayrulewidth} 
\textbf{Input} $S$: Al and Tara can't agree what to watch on TV. Al wants to watch soaps. 
Tara wants to watch the news. They play Paper, Rock, Scissors to decide. Tara wins and gets to watch what she wants. \newline
$\mathbf{\alpha}'$: Al and Tara have two T.V.s \\ \Xhline{1\arrayrulewidth}
\textbf{Output} Al and Tara can't agree what to watch on TV. Al wants to watch soaps. 
Tara wants to watch the \colorbox{pink}{tv}. They play Paper, Rock, Scissors to decide. {Tara wins and gets to watch what she wants.} \\
\Xhline{4\arrayrulewidth}
2. Illogical revised story  \\ \Xhline{4\arrayrulewidth} 
\textbf{Input} $S$: Rich was a musician. He made a few hit songs. 
Rich had a lot of fans who cared about him. He developed cancer. 
Eventually the cancer took his life and rich was gone. \newline
$\alpha'$: Rich was unknown. 
\\ \Xhline{1\arrayrulewidth}
\textbf{Output} Rich was a musician. He made a few hit songs. 
Rich had \colorbox{pink}{no fans. He became ill and lost his life. Eventually Rich recovered} \colorbox{pink}{and began his recovery.} \\
\Xhline{4\arrayrulewidth}
3. Contradiction  \\ 
\Xhline{4\arrayrulewidth} 
\textbf{Input} $S$: There was a large car wreck about two miles from the office.
I was waiting for a co worker to relieve me.
However, he was very late and I had no idea why.
He called me a few minutes later.
Apparently he was stuck in traffic because of the wreck. \newline
$\alpha'$: \hl{His co worker did not have access to a phone.}
\\ \Xhline{1\arrayrulewidth}
\textbf{Output} There was a large car wreck about two miles from the office.
I was waiting for a co worker to relieve me.
However, he was very late and I had no idea why.
\hl{He called me on the phone} \colorbox{pink}{a few minutes later but I had no idea why.} Apparently he was stuck in traffic because of the wreck. 
\\\Xhline{4\arrayrulewidth}
4. Input state not entailed  \\ 
\Xhline{4\arrayrulewidth} 
\textbf{Input} $S$: Frank was very skilled at shaping sculptures.
 His artwork started creating quite a buzz.
 A plain looking man approached Frank with a job.
 He wanted him to design the face for an adult themed product.
 Frank was about to decline until he saw the amount of the check.\newline
$\alpha'$: The check was for a very small amount of money.
\\ \Xhline{1\arrayrulewidth}
\textbf{Output} Frank was very skilled at shaping sculptures.
 His artwork started creating quite a buzz.
 A plain looking man approached Frank with a job.
 He wanted him to design the face for an adult themed product.
 Frank was about to decline \colorbox{pink}{but didn't have much money}.
\\\Xhline{4\arrayrulewidth}
\end{tabular}
\caption{\small{Representative examples of top failure categories for the \textbf{\storyrev{}} task. Parts of the story that are changed are highlighted in \colorbox{pink}{pink}, and the contradictory elements are highlighted in \hl{yellow}.}}
\label{Table:examples_1}
\end{figure}
\begin{figure}[t]
\scriptsize
\begin{tabular}{p{7cm}}
\Xhline{4\arrayrulewidth} 
1. Irrelevant states  
\\ \Xhline{4\arrayrulewidth} 
\textbf{Input} $S$: Pat was looking for an interesting place to vacation. He searched various catalogs. Eventually he found the perfect destination. There he stayed in a small hut. It was one of the greatest experiences of his life.
$S'$: Pat was looking for a relaxing place to vacation. He searched various catalogs. Eventually he found the perfect destination. There he stayed in a luxury resort.
It was one of the greatest experiences of his life. 
\\ \Xhline{1\arrayrulewidth}
\textbf{Output}  $\alpha$: Pat likes to be independent.	$\alpha'$: Pat likes to be in a big city.
\\ \Xhline{4\arrayrulewidth}
2. Illogical state change  \\ \Xhline{4\arrayrulewidth} 
\textbf{Input} $S$: Tom spent a long time writing an essay. He saved it on his memory card. When Tom went to print it the essay wasn't there. Tom had to go back to his computer. Luckily he was able to use a backup copy he made.
$S'$: Tom spent a long time writing an essay. He saved it on his memory card. When Tom went to print it the essay wasn't there. Tom had to go back to his computer. Unfortunately, Tom forgot to make a backup copy of the essay.
\\ \Xhline{1\arrayrulewidth}
\textbf{Output} $\alpha$: Tom is a good memory writer. $\alpha'$: Tom is a bad memory writer.
\\ \Xhline{4\arrayrulewidth}
3. Contradiction  \\ 
\Xhline{4\arrayrulewidth} 
\textbf{Input} $S$: Rita accidentally left her smartphone on the plane. She felt terrible about this and knew her parents would be mad. Nobody ever turned in the phone, leaving her disappointed with people. \hl{Rita had to work many hours to buy a new phone as a replacement.} She is significantly more careful with her belongings now. 
$S'$: Rita accidentally left her smartphone on the plane.
 She felt terrible about this and knew her parents would be mad.
 Nobody ever turned in the phone, leaving her disappointed with people.
 Rita quickly bought a new phone as a replacement.
 She is significantly more careful with her belongings now.
\\ \Xhline{1\arrayrulewidth}
\textbf{Output} $\alpha$: \hl{Rita was unemployed} $\alpha'$: Rita was rich
\\\Xhline{4\arrayrulewidth}
\end{tabular}
\caption{\small{Examples of top failure categories for the \textbf{\statechange{}} task. For the contradiction category, the conflicting parts are \hl{highlighted}.}}
\label{Table:examples_2}
\end{figure}

\noindent\textbf{\mcq{}: }
 We analyze model performance on different types of entity states following the categorization from ~\citet{bhagavatula2019abductive}. We expand their spatial category to a broader set of physical attributes of entities (e.g. weight, temperature, location, etc.), and include a new Societal category to capture social constructs and norms. %
 Even though multiple categories may apply to a state, to simplify our analysis we only use the most relevant category for each state.

In particular, we categorize each instance into one of the following: (i) Societal: knowledge about societal constructs such as relationship (\emph{Jake is not married, I have 5 brothers}), norms (\emph{John is not socially aware}), etc. (ii) Emotional/Psychological: knowledge about emotions (\emph{John felt embarrassed, John hated Jake}), beliefs (\emph{Jake believed in ghosts}), etc. (iii) Physical: Knowledge about physical attributes of entities (\emph{Jake was in his school, the rock was very heavy, the coffee was hot, etc.}). Table~\ref{Table:task8-error} breaks down the overall performance of models across different categories. Models significantly under-perform on the societal category compared to the other two. 
In addition to the difficulty of modeling societal knowledge, we find that relatively more number of instances in this category require numerical commonsense, which adds additional complexity for the models.
Physical commonsense is a broad category and its instances thus tend to cover a broad range of physical knowledge which could contribute to the difficulty of these instances. Emotional category has the best model performance since the inferred state include strong lexical indicators of emotions and feelings, similar to the observations in \citet{bhagavatula2019abductive}.

The proposed generative tasks can have multiple correct outputs, each using a different set of commonsense knowledge. This makes it difficult to associate a unique knowledge category for the task instance. Therefore we manually analyze the outputs of the best performing model (T5-large) and identified common types of generation errors made by the model on the task.

\noindent\textbf{\storyrev{}: }
The model output is correct for $\sim 58\%$ cases and incorrect for $\sim42\%$.
On analyzing the incorrect output, we found four main categories of error that we list in Table~\ref{Table:task6-error}.
The "illogical revised story" occurs when models produced revised stories that are logically incoherent (30\% of errors). Generating logically coherent long text is still a challenging task for models, and to a certain extent can be attributed to their tendency to forget attributes of specific entities \cite{Welleck2018DialogueNL}, ignore previously inferred facts \cite{sinha-etal-2019-clutrr} and background information, or contradict previous statements \cite{brown2020language}.
Moreover, $20.5\%$ of the revised stories are categorized as \textit{contradiction} as they clearly contradict the input counterfactual state. This corroborates previous findings on the challenges in reasoning about contradictions and negations~\cite{hossain-etal-2020-analysis}.
Models also struggle to keep the changes relevant to the task criteria of the input state, which should be inferable from the revised story but not directly mentioned in it.
They sometimes make \textit{Irrelevant changes} ($27.7\%$ of errors) where they revise parts of the story that  are not affected by the input counterfactual state. 
Other times they make revision that are inconsistent with the input counterfactual state (\textit{Input state not entailed}, $20.5\%$) or the input \textit{State is explicit in the revision} ($1.2\%$), both of which do not meet the primary task requirements.

%
Figure~\ref{Table:examples_1} shows examples of the biggest error categories for the task.

\noindent\textbf{\statechange{}:}
The model is correct for 54.5\% of cases and fails for 45.5\% when generating state changes. 
Table~\ref{Table:task7-error} shows the main error categories. 
While the model learns to generate both the $\alpha$ and $\alpha'$ states about the same entity, it makes many types of logical errors.
\textit{Contradictions} ($37.4\%$ of errors) are when a generated state is contradicted by its story, either directly or by deduction. 
\textit{Illogical State Changes} ($13.2\%$) are those where the generated states and input stories were topically related, but the states were simply illogical  nonsensical.
Both types of errors can be attributed to the challenges associated with making the relevant state inference, generating logically coherent text and reasoning about contradictions and negations.
\textit{Irrelevant States} ($35.2\%$) are those where at least one of the generated states has no connection to its story.
The error categories of \textit{State Reversed} ($4.4\%$), \textit{No change in state} inferability ($4.4\%$), \textit{State is directly stated in the story} ($4.4\%$) and outputs are \textit{Actions instead of states} ($1.1\%$) are due to models' inability to correctly understand the task constraints.
Figure~\ref{Table:examples_2} shows example of the major error categories for the task.

\subsection{Interactive Feedback with LLMs}
Based on the error analysis for the tasks performed above, the majority of the error categories can be attributed to the model's inability to maintain factual and logical consistency in the generated output. For the \mcq{} task, the lack of consistency is further demonstrated by the low contrastive accuracy on the task.
Conversation-based LLMs such as ChatGPT~\cite{chatgpt} or LaMDA~\cite{lamda}, have been shown to have both knowledge at the scale of LLMs such as GPT3 and an ability to incorporate human feedback for NLU tasks. 
These capabilities may enable them to leverage feedback about inconsistencies (if detected) in the initially generated output to correct these inconsistencies in the subsequent generations.
However, when the task is to be performed at scale, the feedback that guides the model to the correct output needs to be automatically generated instead of a human guiding the model. As such, this type of model presents a fruitful and challenging research direction to address some of the issues and further improve performance on the tasks.
\section{Conclusion}
In this work, we introduced a new resource, \dataset{}, that captures unstated commonsense knowledge required to understand and reason about participant states in a narrative. \dataset{} opens the door to developing more complex reasoning abilities, especially those that require access to implicit information. We described three \dataset{} reasoning tasks, one classification and two generation, that test for different aspects of state-based reasoning. This work shows that with careful crowdsourcing and contrastive design we can obtain a high-quality dataset that can be used to evaluate deeper reasoners. Benchmarking results suggest that \dataset{} tasks are not within the reach of current large sized models, as of yet, and encourages future research in modeling commonsense knowledge with states. 



\section*{Acknowledgments}
We would like to thank the anonymous reviewers for their comments, questions, and suggestions. %
This material is also based on research that is in part supported by the NSF, Grant No. 2007290, Army Research Laboratory, Grant No. W911NF2120076, and by the Air Force Research Laboratory (AFRL), DARPA, for the KAIROS program under agreement number FA8750-19-2-1003. The U.S. Government is authorized to reproduce and distribute reprints for Governmental purposes notwithstanding any copyright notation thereon. The views and conclusions contained herein are those of the authors and should not be interpreted as necessarily representing the official policies or endorsements, either express or implied, of the Air Force Research Laboratory (AFRL), DARPA, or the U.S. Government. %
This material is based in part upon work supported by the National Science Foundation under Grant No. IIS-2024878. %
\bibliography{tacl2021}
\bibliographystyle{acl_natbib}

\iftaclpubformat

\onecolumn






  
\fi

\end{document}